\pdfoutput=1

\documentclass[11pt]{article}

\usepackage{amsmath,amsfonts,bm}

\def\eqref#1{equation~\ref{#1}}

\def\1{\bm{1}}

\DeclareMathAlphabet{\mathsfit}{\encodingdefault}{\sfdefault}{m}{sl}
\SetMathAlphabet{\mathsfit}{bold}{\encodingdefault}{\sfdefault}{bx}{n}

\usepackage{adjustbox}
\usepackage{algorithm}
\usepackage{algorithmicx}
\usepackage[noend]{algpseudocode}
\usepackage{amsmath,amssymb,amsfonts,amsthm,mathtools}
\usepackage{bm,bbm}
\usepackage{duckuments}
\usepackage{enumitem}
\usepackage{graphicx}
\usepackage{float}
\usepackage[nice]{nicefrac}
\usepackage[multiple]{footmisc}
\usepackage{caption,subcaption}
\usepackage{cprotect}  %
\usepackage{textcomp}
\usepackage{stfloats}
\usepackage[colorlinks=true,linkcolor=Blue9,citecolor=Blue9]{hyperref}
\usepackage{cleveref}
\usepackage{listings}
\usepackage{lipsum}
\usepackage{longtable,tabularx,booktabs,wrapfig}
\usepackage{multirow}
\usepackage{siunitx}
\usepackage{url}
\usepackage{xcolor}
\usepackage{xspace}
\usepackage[utf8]{inputenc}
\usepackage{pgfplots}
\usepackage{tcolorbox,tabularray}
\usepackage{placeins}
\usepackage{colortbl}
\DeclareUnicodeCharacter{2212}{−}
\usepgfplotslibrary{groupplots,dateplot}
\usetikzlibrary{patterns,shapes.arrows}
\pgfplotsset{compat=newest}

\usepackage{array}
\newcolumntype{R}[2]{%
    >{\adjustbox{angle=#1,lap=\width-(#2)}\bgroup}%
    l%
    <{\egroup}%
}

\newcolumntype{Y}{>{\centering\arraybackslash}X}

\DeclarePairedDelimiterX{\infdivx}[2]{(}{)}{%
  #1\;\delimsize\|\;#2%
}
\newcommand{\infdiv}{D_{\text{KL}}\infdivx}

\definecolor{Gray}{gray}{0.9}
\definecolor{Blue9}{rgb}{0.098,0.3,0.9}
\definecolor{Red7}{rgb}{0.941, 0.243, 0.243}
\definecolor{Green7}{RGB}{55, 178, 77}
\definecolor{BrickRed}{rgb}{0.6,0,0}
\definecolor{RoyalBlue}{rgb}{0,0,0.8}
\definecolor{Tdgreen}{rgb}{0,0.4,0.7}
\definecolor{CYRed}{RGB}{228, 0, 43}
\definecolor{CYPurple}{RGB}{215, 153, 93}

\usepackage[final]{acl}

\usepackage{times}
\usepackage{latexsym}

\usepackage[T1]{fontenc}

\usepackage[utf8]{inputenc}

\usepackage{microtype}

\usepackage{inconsolata}

\usepackage{graphicx}

\title{Margin Matching Preference Optimization:\\Enhanced Model Alignment with Granular Feedback}

\author{
    \textbf{Kyuyoung Kim}$^{1*}$,  %
    \textbf{Ah Jeong Seo}$^{1*}$,  %
    \textbf{Hao Liu}$^{2}$,  %
    \textbf{Jinwoo Shin}$^{1}$,  %
    \textbf{Kimin Lee}$^{1}$  %
\\
    \textsuperscript{1}KAIST,
    \textsuperscript{2}UC Berkeley
}

\begin{document}
\maketitle
\def\thefootnote{*}\footnotetext{Equal contribution}\def\thefootnote{\arabic{footnote}}

\begin{abstract}
Large language models (LLMs) fine-tuned with alignment techniques, such as reinforcement learning from human feedback, have been instrumental in developing some of the most capable AI systems to date.
Despite their success, existing methods typically rely on simple binary labels, such as those indicating preferred outputs in pairwise preferences, which fail to capture the subtle differences in relative quality between pairs.
To address this limitation, we introduce an approach called Margin Matching Preference Optimization (MMPO), which incorporates relative quality margins into optimization, leading to improved LLM policies and reward models.
Specifically, given quality margins in pairwise preferences, we design soft target probabilities based on the Bradley-Terry model, which are then used to train models with the standard cross-entropy objective.
Experiments with both human and AI feedback data demonstrate that MMPO consistently outperforms baseline methods, often by a substantial margin, on popular benchmarks including MT-bench and RewardBench.
Notably, the 7B model trained with MMPO achieves state-of-the-art performance on RewardBench as of June 2024, outperforming other models of the same scale.
Our analysis also shows that MMPO is more robust to overfitting, leading to better-calibrated models.
\end{abstract}

\section{Introduction}

Large language models (LLMs) trained on internet-scale data have demonstrated remarkable instruction-following and generalization capabilities, leading to their widespread adoption across natural language processing (NLP) tasks~\citep{brown2020language,palm2023,penedo2023refinedweb,chung2024scaling}.
Pre-trained on a large, general corpus of text, LLMs acquire broad knowledge about the world with language understanding and reasoning abilities~\citep{radford2019language,wei2022emergent,wei2022chain,kojima2022large}.
To adapt a pre-trained LLM to a downstream task, the model is typically fine-tuned on demonstrations of the desired output for the task.
However, providing high-quality demonstrations is often more costly than simply evaluating model outputs.
To address this, reinforcement learning from human feedback (RLHF;~\citealt{christiano2017deep,stiennon2020learning}) and AI feedback (RLAIF;~\citealt{cai,lee2023rlaif}) utilize feedback on diverse outputs to optimize LLMs, aligning the responses more closely with human intent.
RLHF has been instrumental in developing some of the most capable AI systems to date~\citep{gpt_4,team2023gemini,anthropic2024claude}.

Feedback-based alignment leverage human or AI feedback data to fine-tune generative models to better align their outputs with human intent.
Reward-based methods such as RLHF use feedback to learn a surrogate reward function, which is subsequently used in fine-tuning models via, e.g., reinforcement learning (RL).
In contrast, reward-free methods, such as direct preference optimization (DPO;~\citealt{rafailov2024direct}), bypass explicit reward modeling and directly fine-tune models using the data.
Feedback is commonly in the form of \textit{pairwise} preferences, where responses to a given input are compared in pairs (e.g., $y_1 \succ y_2$ for input $x$), and labels indicating the preferred one of the two are collected.
More recent methods, such as Kahneman-Tversky optimization (KTO;~\citealt{ethayarajh2024kto}), also utilize simpler binary feedback indicating whether or not a response to an input is desirable.
While these alignment methods have shown to be more effective than applying SFT alone, existing methods rely only on binary labels indicating either the preferred output in a pair or the desirability of a single output, forgoing the opportunity to incorporate more granular feedback signals into learning.

\begin{figure*}[t]
    \centering
    \includegraphics[width=0.95\linewidth]{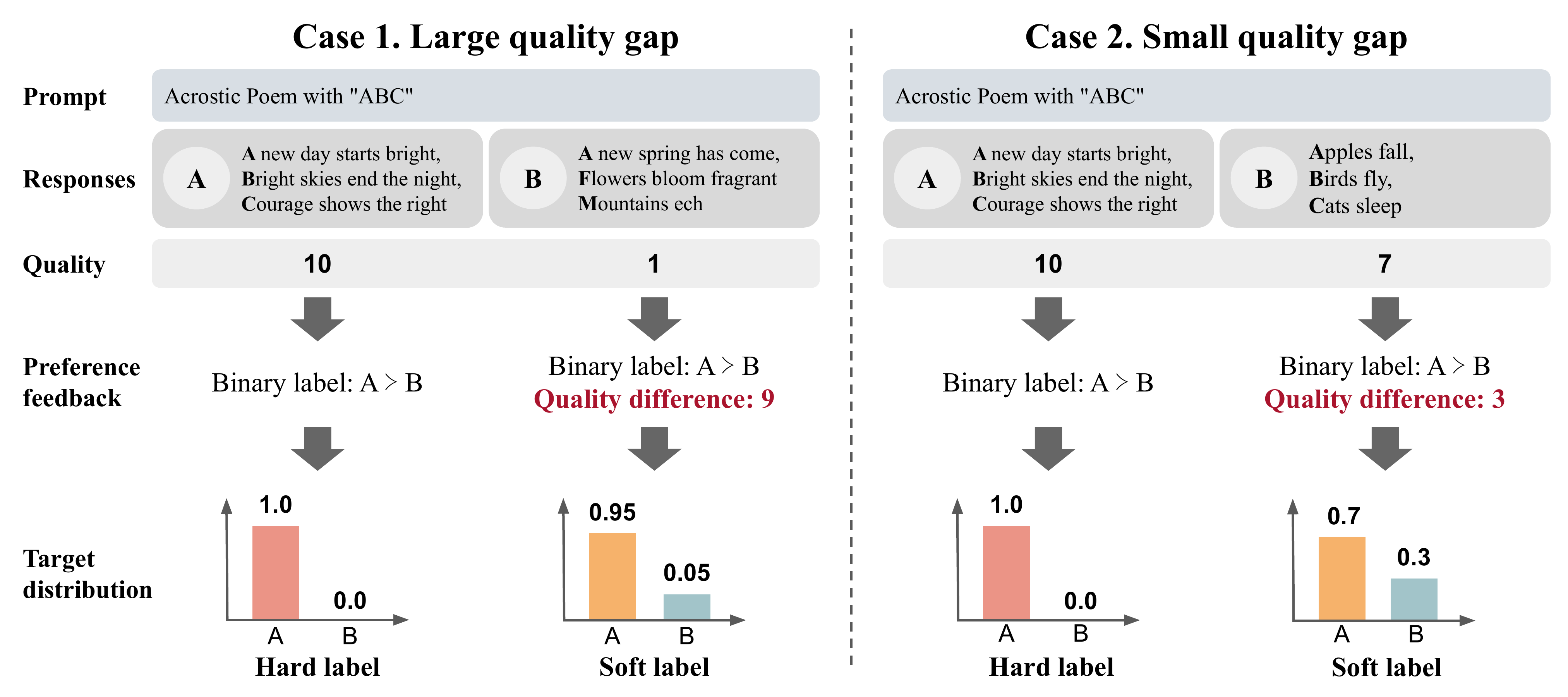}
    \caption{The quality gap between response pairs in pairwise preference data often varies significantly. By incorporating granular feedback into optimization, MMPO enhances model performance, robustness, and calibration.}
    \vspace{-0.15in}
    \label{fig:concept}
\end{figure*}

In this work, we introduce Margin Matching Preference Optimization (MMPO), a simple yet effective generalization of common alignment methods that integrates granular feedback signals into optimization, enabling models to capture subtle preferences in the feedback data (see Figure~\ref{fig:concept}).
Such granular feedback can come from human annotators providing detailed ratings, such as Likert scale scores, or from AI models, which are increasingly used for automatic evaluation of model responses.
The MMPO objective utilizes per-sample target preference probabilities, designed based on the quality margin between output pairs.
Since reward modeling and DPO, as well as similar methods based on pairwise preferences, use equivalent cross-entropy loss, the approach extends naturally to both.
By incorporating target probabilities that reflect the quality margin of each output pair, models are trained to better account for the specifics of each feedback sample.
For instance, when the quality margin is large, models trained with MMPO would assign significantly higher scores to preferred outputs, while assigning similar scores when the margin is narrow.
The idea easily extends to methods that rely on alternative forms of feedback.

Our main contributions are as follows:
\begin{itemize}[leftmargin=4mm]
    \item We present Margin Matching Preference Optimization (MMPO), a simple generalization of alignment methods that utilizes granular feedback signals to enhance model alignment.
    \item Our empirical results on both human and AI feedback demonstrate that MMPO outperforms baseline methods on MT-bench~\citep{zheng2024judging}, a popular benchmark for evaluating model generation quality, achieving improvements of up to 11\%.
    \item Evaluation on RewardBench~\citep{lambert2024rewardbench}, a benchmark designed to assess models' capabilities as reward models, shows that the 7B model trained with MMPO achieves state-of-the-art performance, outperforming competing models at the same scale, as of June 2024.
    \item Our analysis shows that MMPO is more robust to overfitting on feedback data, resulting in well-calibrated models that better generalize to prompts unseen during fine-tuning.
\end{itemize}

\section{Preliminaries}
\label{sec:prelim}

Model alignment using feedback generally involves 1) supervised fine-tuning (SFT) of pre-trained models and 2) aligning the models based on human or AI feedback data.
In the SFT phase, a pre-trained LLM is fine-tuned using supervised learning on task-specific demonstrations, such as human-written summaries in the case of a summarization task.
In the alignment phase, the model is further fine-tuned to generate outputs that align with the preferences reflected in the feedback data.
Below, we review several popular approaches.

\paragraph{RLHF.}
RLHF methods are reward-based approaches that first learn a reward function from feedback data, which is then used to provide training signals to the language model in RL fine-tuning.
Given a dataset $\mathcal{D}$ of pairwise preferences $(x, y_w, y_l)$, where $x$ is an input and $y_w$ and $y_l$ are the preferred and dispreferred outputs, respectively, the reward function is trained to assign a higher score to $y_w$ than to $y_l$.
Specifically, we model human preference probabilities using the Bradley-Terry model~\citep{bradley1952rank}, which defines the probability as the sigmoid function applied to the difference in rewards given by the reward function $r_\phi$:
\begin{equation*}
    \hat{p}(y_w \succ y_l \mid x)
    = \sigma( r_\phi(x, y_w) - r_\phi(x, y_l)),
\end{equation*}
where $\sigma$ is the sigmoid function.
We optimize the parameters of $r_\phi$ by minimizing the following cross-entropy loss on the feedback data
\begin{equation}
    \mathcal{L} = -\mathbb{E}_{(x, y_w, y_l) \sim \mathcal{D}} [\log \sigma(r_\phi(x, y_w) - r_\phi(x, y_l))].
\label{eq:xentropy}
\end{equation}

Following reward learning, we train the language model policy $\pi_\theta$ to maximize the learned reward $r_\phi$ with a constraint that limits excessive deviation from a reference policy $\pi_{\text{ref}}$.
Specifically, we use a policy gradient method, such as Proximal Policy Optimization (PPO;~\citealt{schulman2017proximal}), to maximize the following KL-constrained objective:
\begin{equation}
\begin{aligned}
    \max_{\pi_\theta}\, & \mathbb{E}_{x \sim \mathcal{D},y \sim \pi_\theta} \left[ r_\phi(x, y) \right] \\
    & -  \beta \infdiv{\pi_\theta(y \mid x)}{\pi_{\text{ref}}(y \mid x)} ,
\label{eq:kl-const-opt}
\end{aligned}
\end{equation}
where $\beta$ is a parameter controlling the strength of the constraint.

\paragraph{DPO.} 
An alternative to reward-based approaches is DPO~\cite{rafailov2024direct}, which bypasses both explicit reward modeling and RL-based training.
Instead, DPO leverages an analytical relationship between the reward function and the optimal solution to the KL-constrained optimization objective in Eq.~\ref{eq:kl-const-opt} to derive the following loss:
\begin{equation*}
\begin{aligned}
    & \mathcal{L}_{\text{DPO}} =-\mathbb{E}_{(x, y_w, y_l) \sim \mathcal{D}} \biggr[ \\ & \log \sigma \Bigr(\beta \log \frac{\pi_\theta(y_w \mid x)}{\pi_{\text{ref}}(y_w \mid x)} - \beta \log \frac{\pi_\theta(y_l \mid x)}{\pi_{\text{ref}}(y_l \mid x)} \Bigr) \biggr],
\end{aligned}
\end{equation*}
where $\pi_\theta$ is the language model policy being trained, and $\pi_{\text{ref}}$ denotes a reference policy.
DPO has recently gained popularity because it allows maximum likelihood training of language model policies, which typically requires significantly less computational resources than RL.

\section{Margin Matching Preference Optimization}

Regardless of whether the feedback consists of pairwise preferences or indicators of desirability, the feedback label is generally binary.
For pairwise preferences, we collect a binary label that identifies the preferred output from a pair.
In the case of binary feedback, the label indicates whether or not an output is desirable for a given input.
However, more detailed feedback is often available, particularly with the increasing use of LLMs and other AI models as annotators~\citep{touvron2023llama,cui2023ultrafeedback}.
Motivated by this observation, we propose a simple generalization of common optimization objectives used in feedback learning methods, which results in improved reward models and language model policies.
The main idea is to design per-sample target preference probabilities $p(y_w \succ y_l \mid x)$ for methods such as DPO that rely on pairwise preferences, and to adjust the loss weight for each $(x, y)$ based on the desirability of $y$ for $x$ in methods such as KTO that are based on binary feedback.

\subsection{Limitations of binary labels}

The implicit assumption in Eq.~\ref{eq:xentropy} that the target preference probability $p(y_w \succ y_l \mid x)$ equals 1 for every sample in the feedback dataset leads to several limitations.
First, it overlooks the fact that the Bradley-Terry model defines preference probability in terms of the difference in rewards between the output pairs, i.e., $p(y_w \succ y_l \mid x) = \sigma(r(x, y_w) - r(x, y_l))$.
Consequently, if $y_w$ is only marginally preferred to $y_l$, the preference probability that more accurately captures this subtlety is likely to be far less than 1.
Second, setting the target probability to 1 makes the optimization prone to overfitting, as this is only attainable when $r(x, y_w) - r(x, y_l) = \infty$.
This latter point has been analyzed in the context of DPO~\citep{azar2024general}, but the issue applies to reward modeling as well.

\begin{wrapfigure}{r}{3.5cm}
    \centering
    \vspace{-0.15in}
    \includegraphics[width=\linewidth]{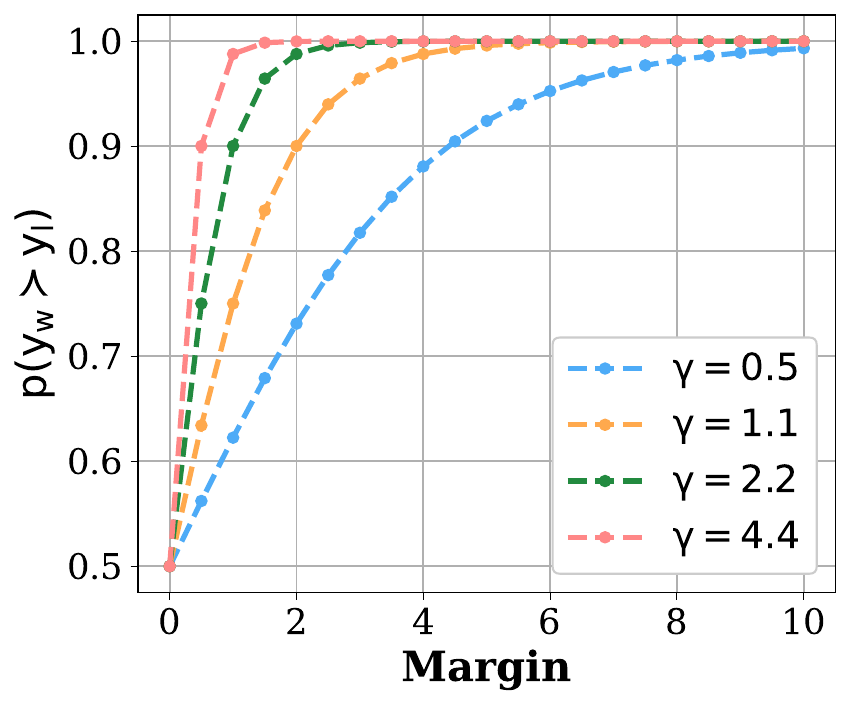}
    \vspace{-0.25in}
    \caption{Bradley-Terry model's preference probabilities with varying $\gamma$.}
    \vspace{-0.15in}
    \label{fig:bt-example}
\end{wrapfigure}

In case more detailed information on individual feedback samples is available, such as the relative difference in quality between $y_w$ and $y_l$ in pairwise preferences, we can formulate optimization objectives that more accurately capture per-sample characteristics and are more robust to overfitting.
This information may come from human annotators providing more fine-grained ratings~\citep{touvron2023llama} or from LLM judges~\citep{zheng2024judging} assigning scores to individual samples.
For the remainder, we omit the dependence on input $x$ for simplicity.

\subsection{Generalized preference optimization}
\label{subsec:mmpo}
Given the quality difference between $y_w$ and $y_l$, denoted as $m(y_w, y_l) < \infty$, we use the fact that the Bradley-Terry model depends only on the difference in rewards to design the target preference probability based on the quality margin as follows:
\begin{equation*}
\begin{aligned}
    p(y_w \succ y_l) = \sigma(r(y_w) - r(y_l)) = \sigma(\gamma m(y_w, y_l)),
\end{aligned}
\end{equation*}
where $\gamma$ is a scaling parameter commonly referred to as the rationality coefficient.
As $\gamma \rightarrow \infty$, preferences becomes perfectly rational and deterministic, always favoring the choice with the higher reward.
In contrast, when $\gamma = 0$, preferences become uniformly random, resulting in an equal likelihood of favoring either choice regardless of the underlying rewards.
Figure~\ref{fig:bt-example} illustrates how the change in preference probability derived from the Bradley-Terry model varies across different values of $\gamma$ for score differences ranging from 0 to 10.

Since $m(y_w, y_l)$ is finite, $p(y_w \succ y_l)$ is less than 1, which leads to the more general binary cross-entropy loss that is also less susceptible to overfitting the feedback data,
\begin{equation*}
\begin{aligned}
    \mathcal{L} =& -\mathbb{E}_{(y_w, y_l) \sim \mathcal{D}} \Bigr[ \\&p(y_w \succ y_l) \log \sigma(\hat{r}(y_w) - \hat{r}(y_l)) \\
    & +\, (1 - p(y_w \succ y_l)) \log \sigma(\hat{r}(y_l) - \hat{r}(y_w)) \Bigr].
\end{aligned}
\end{equation*}
For reward modeling, $\hat{r}$ is simply the parameterized reward function, $r_\phi$.
For DPO, $\hat{r}$ is the implicit reward defined by the language model policy $\pi_\theta$ and reference policy $\pi_{\text{ref}}$, resulting in the loss
\begin{equation*}
\scalebox{0.72}{%
$\begin{aligned}
    &\mathcal{L}_{\text{DPO}} = -\mathbb{E}_{(y_w, y_l) \sim \mathcal{D}} \biggr[ \\& \sigma(\gamma m(y_w, y_l)) \, \log \sigma\Bigr(\beta \log \frac{\pi_\theta(y_w)}{\pi_{\text{ref}}(y_w)} - \beta \log \frac{\pi_\theta(y_l)}{\pi_{\text{ref}}(y_l)} \Bigr) \\
    & +\, (1 - \sigma(\gamma m(y_w, y_l))) \, \log \sigma\Bigr(\beta \log \frac{\pi_\theta(y_l)}{\pi_{\text{ref}}(y_l)} - \beta \log \frac{\pi_\theta(y_w)}{\pi_{\text{ref}}(y_w)} \Bigr) \biggr].
\end{aligned}$}
\end{equation*}

Intuitively, we train models to align with per-sample preference probabilities, which are determined by the quality margin between each pair of $y_w$ and $y_l$.
If $y_w$ is of significantly better quality than $y_l$, the target preference probability that the models are trained to fit would be close to 1.
Conversely, if both outputs are of comparable quality, the probability would be closer to 0.5.
In other words, the models are trained to capture these per-sample subtleties.
This core idea can also be easily extended to other forms of feedback, which we discuss further in Appendix~\ref{appendix:extension}.

\section{Experiments}

In this section, we evaluate MMPO by assessing the performance of the language model policies and reward models trained using the method on popular benchmarks.
We utilize both human and AI feedback data, presenting benchmark performance along with an analysis of model calibration and the method's robustness to overfitting.

\subsection{Setup}
\label{subsec:setup}

\paragraph{Supervised fine-tuning.}
We conduct our experiments using the Gemma models, both the 2B and 7B variants, and the Llama 3 model at the 8B scale, which are state-of-the-art open LLMs at their respective scales~\citep{team2024gemma,dubey2024llama}.
We first apply supervised fine-tuning (SFT) to the pre-trained models on UltraChat~\citep{ding2023enhancing}, a dialogue dataset that has been used to produce strong chat models such as UltraLM~\citep{ding2023enhancing}.
The dataset comprises multi-turn dialogues across 30 topics and includes 20 types of text materials generated using ChatGPT.
In particular, we use the refined version of the dataset, with various filters applied to remove undesirable responses, consisting of 200k samples also used in training the recent Zephyr model~\citep{tunstall2023zephyr}.
The SFT models are used for both direct alignment with preference data and for reward modeling.
Additional experimental details can be found in Appendix~\ref{appendix:training}.

\paragraph{Feedback datasets.}
We evaluate alignment methods using both human and AI feedback to assess their performance across various types of feedback data.
UltraFeedback~\citep{cui2023ultrafeedback} is a dataset consisting of 64k prompts and pairs of responses generated using a diverse set of LLMs.
Each response is rated on a scale of 1 to 10 by GPT-4~\citep{achiam2023gpt}, based on criteria such as instruction-following and helpfulness.
Given the ratings for individual responses, we use them to compute the quality margin for each pair.
We then compute the target preference probability following the Bradley-Terry model, with the scaling parameter $\gamma$ tuned based on validation accuracy.

To also experiment with human feedback, we use the SHP dataset~\citep{ethayarajh2022understanding}, which consists of human preferences for responses to Reddit posts across 18 subject areas.
The dataset provides scores for each response based on the number of positive and negative votes received from users, serving as a proxy for the relative quality of the responses.
We compute the target preference probabilities in a similar manner, using the scores derived from the net positive number of votes.
Given the large size of the original dataset, we construct a sample of 55k responses, which is comparable in size to UltraFeedback.
While training exclusively on preferences with significant score differences has been shown to result in better performing models~\citep{ethayarajh2022understanding}, we sample uniformly across score differences to evaluate the methods on diverse quality margins.
Further details on dataset sampling can be found in Appendix~\ref{appendix:training}.

\begin{figure*}[t]
    \centering
    \hfill
    \begin{minipage}{0.47\linewidth}
        \captionof{table}{MT-bench results for models trained with MMPO and DPO. The results for other open and proprietary models are from the official leaderboard.}
        \centering
        \small

\begin{tabularx}{\textwidth}{lc|*{2}{Y}}
    \toprule
    Model & Size & UF & SHP \\
    \midrule
    Gemma-SFT & 2B & 4.73 & 4.73 \\
    Gemma-DPO & 2B & \underline{6.09} & \underline{5.13} \\
    Gemma-MMPO & 2B & \textbf{6.10} & \textbf{5.57} \\
    \midrule
    Gemma-SFT & 7B & 6.84 & 6.84 \\
    Gemma-DPO & 7B & \underline{7.40} & \underline{6.49} \\
    Gemma-MMPO & 7B & \textbf{7.53} & \textbf{7.23} \\
    \midrule
    \midrule
    Gemma-IT & 7B & \multicolumn{2}{c}{6.26} \\
    Zephyr$-\beta$ & 7B & \multicolumn{2}{c}{7.34} \\
    GPT-3.5-Turbo & - & \multicolumn{2}{c}{7.94} \\
    GPT-4 & - & \multicolumn{2}{c}{8.99} \\
    \bottomrule
\end{tabularx}

        \label{table:benchmark-mtbench}
    \end{minipage}
    \hfill
    \begin{minipage}{0.47\linewidth}
        \centering
        \includegraphics[width=1.0\linewidth]{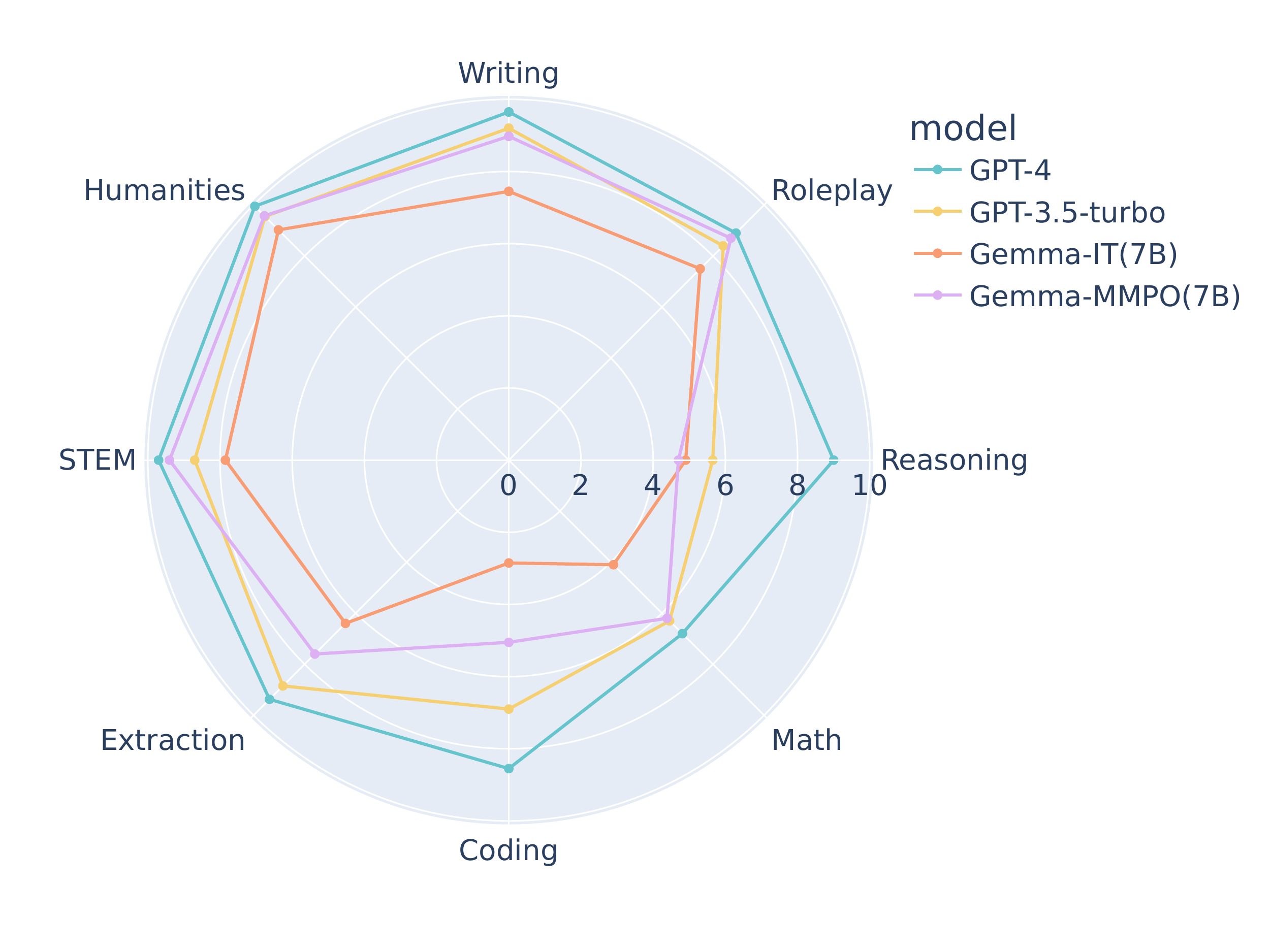}
        \vspace{-0.45in}
        \caption{MT-bench results categorized by the eight domains. The MMPO model outperforms Gemma-IT and is competitive with GPT-3.5 in multiple domains.}
        \label{fig:mtbench-fig}
    \end{minipage}
    \hfill
    \vspace{-0.15in}
\end{figure*}

\paragraph{Evaluation.}
Our main evaluations are based on MT-bench~\citep{zheng2024judging}, a multi-turn chat benchmark consisting of 160 questions across eight domains.
On this benchmark, models are assessed on their ability to follow instructions and respond coherently over two turns of conversation.
Each of the two responses is evaluated by GPT-4 as a proxy for human judgments on a scale of 1 to 10, with the average score used as the score for that conversation.
The final benchmark score is calculated as the mean score across all 160 conversations.

Additionally, we evaluate models on RewardBench~\citep{lambert2024rewardbench}, a benchmark designed to assess models' capabilities as reward models, i.e., their ability to distinguish between preferred and dispreferred responses to prompts.
This evaluation involves assessing the capability of reward functions, trained as described in Section~\ref{sec:prelim}, to assign higher scores to preferred responses, as well as the ability of language model policies, trained using methods such as DPO, to assign higher implicit rewards for those same preferred responses.
RewardBench contains a diverse set of pairwise preference data to evaluate models across various domains, including chat, safety, and reasoning.
The primary metric for this benchmark is the weighted mean accuracy across all prompts.

\subsection{Direct alignment with MMPO}
\label{subsec:benchmark}

\paragraph{Generation quality.}
Table~\ref{table:benchmark-mtbench} summarizes the MT-bench results for the SFT models and the models fine-tuned on UltraFeedback (UF) and SHP using MMPO and DPO.
Across both synthetic and human feedback data, MMPO consistently produced models that outperform those fine-tuned with DPO.
This performance gap is more significant in the 7B model than in the 2B model, and it is larger when using human feedback data compared to synthetic data.
Notably, the 7B model fine-tuned with DPO on SHP data performs worse than the original 7B SFT model, whereas the 7B model fine-tuned with MMPO shows a clear improvement over both.
This disparity may be due to the fact that human preferences tend to be noisy, underscoring the need to account for per-sample specifics during fine-tuning.
We also evaluated identity preference optimization (IPO;~\citealt{azar2024general}) with the 7B model.
However, it performed significantly worse than DPO, achieving a score of 6.50 when fine-tuned on UltraFeedback.
See Appendix~\ref{appendix:shp-sample} for qualitative examples of model comparison on the SHP dataset.

Figure~\ref{fig:mtbench-fig} shows the MT-bench results for GPT-4, GPT-3.5, the instruction-tuned (IT) 7B model, and the 7B model fine-tuned on UltraFeedback using MMPO, categorized by the eight domains.
The 7B MMPO model outperforms the instruction-tuned model across all domains except reasoning.
Also, while the overall score for the 7B MMPO model is slightly lower than that of GPT-3.5, it matches GPT-3.5's performance in several domains, including humanities and math, and even exceeds it in others, such as STEM and roleplay.

\paragraph{Capability as reward models.}

\begin{table*}[t]
    \caption{RewardBench results for the models fine-tuned with MMPO and DPO on UltraFeedback and for other LLMs from the official leaderboard. The \texttt{Chat} and \texttt{Chat Hard} subsets include open-ended prompts, \texttt{Safety} covers prompts desiged to evaluate the model's ability to avoid harmful content, \texttt{Reason} focuses on prompts that assess coding and reasoning capabilities, and \texttt{Prior Sets} consists of test prompts sampled from datasets such as the Anthropic HH~\citep{bai2022training} and OpenAI's summarization~\citep{stiennon2020learning} datsets.}
    \centering
    \footnotesize
    \begin{adjustbox}{width=0.95\linewidth}
    \begin{tabularx}{\textwidth}{lc|*{1}{Y}|*{6}{Y}}
    \toprule
    Model & Size &\textbf{Avg} & Chat & Chat Hard & Safety & Reason & Prior Sets  \\
    \midrule
    Gemma-DPO & 2B & 59.4 & 95.0 & \textbf{45.6} & 51.9 & 49.6 & 50.1  \\
    Gemma-MMPO & 2B & \textbf{62.3} & \textbf{96.1} & 45.1 & \textbf{52.3} & \textbf{59.8} & \textbf{53.6} \\
    \midrule
    Gemma-DPO & 7B & 73.0 & 96.6 & 59.9 & \textbf{73.7} & 69.0 & 58.3  \\
    Gemma-MMPO & 7B & \textbf{75.6} & \textbf{97.5} & \textbf{62.9} & 71.1 & \textbf{75.0} & \textbf{67.7}  \\
    \midrule
    \midrule
    Zephyr-$\beta$ & 7B & 70.7 & 95.3 & 62.6 & 54.1 & 89.6 & 52.2  \\
    Zephyr-$\alpha$ & 7B & 73.6 & 91.6 & 63.2 & 70.0 & 89.6 & 53.5 \\
    \midrule
    Tulu-2-DPO & 70B & 77.0 & 97.5 & 60.8 & 85.1 & 88.9 & 52.8  \\
    \bottomrule
\end{tabularx}

    \end{adjustbox}
    \label{table:benchmark-rewardbench}
\end{table*}

We also assess the models' capability as reward models, specifically their ability to distinguish between preferred and dispreferred responses across various domains.
For this evaluation, we use RewardBench~\citep{lambert2024rewardbench}, a benchmark consisting of prompts and pairwise preferences designed for such assessments.
Table~\ref{table:benchmark-rewardbench} summarizes the results, with the \texttt{Avg} column showing the overall evaluation score, and the other columns reporting scores for each prompt subset.
The MMPO models consistently outperform the DPO models at both scales across all subsets, with the exception of one case where the results are comparable.
Notably, the performance gap is particularly significant on the \texttt{Reason} and \texttt{Prior Sets} subsets, where the MMPO models substantially outperform the DPO models.
Given that the fine-tuning datasets primarily focus on enhancing chat capabilities, it is notable that the MMPO models achieve superior results on prompt types that differ from those seen during fine-tuning.

Moreover, the MMPO models outperform both Zephyr-$\alpha$ and Zephyr-$\beta$ models, which are two competitive open models at the 7B scale.
As of June 2024, the 7B MMPO model achieves state-of-the-art performance on the RewardBench leaderboard among models at the same scale.
In particular, the 7B MMPO model remains competitive against the much larger Tulu-2-DPO model, which is 10x its size, and even outperforms it on the \texttt{Prior Subsets}.
These results suggest that MMPO, by training models to align with the quality margin of individual feedback samples, results in more calibrated models that better generalize to the types of prompts unseen during fine-tuning.

\paragraph{Calibration analysis.}
\label{subsec:calib}

\begin{figure}[t]
    \centering
    \begin{subfigure}[t]{0.49\linewidth}
        \centering
        \includegraphics[width=\linewidth]{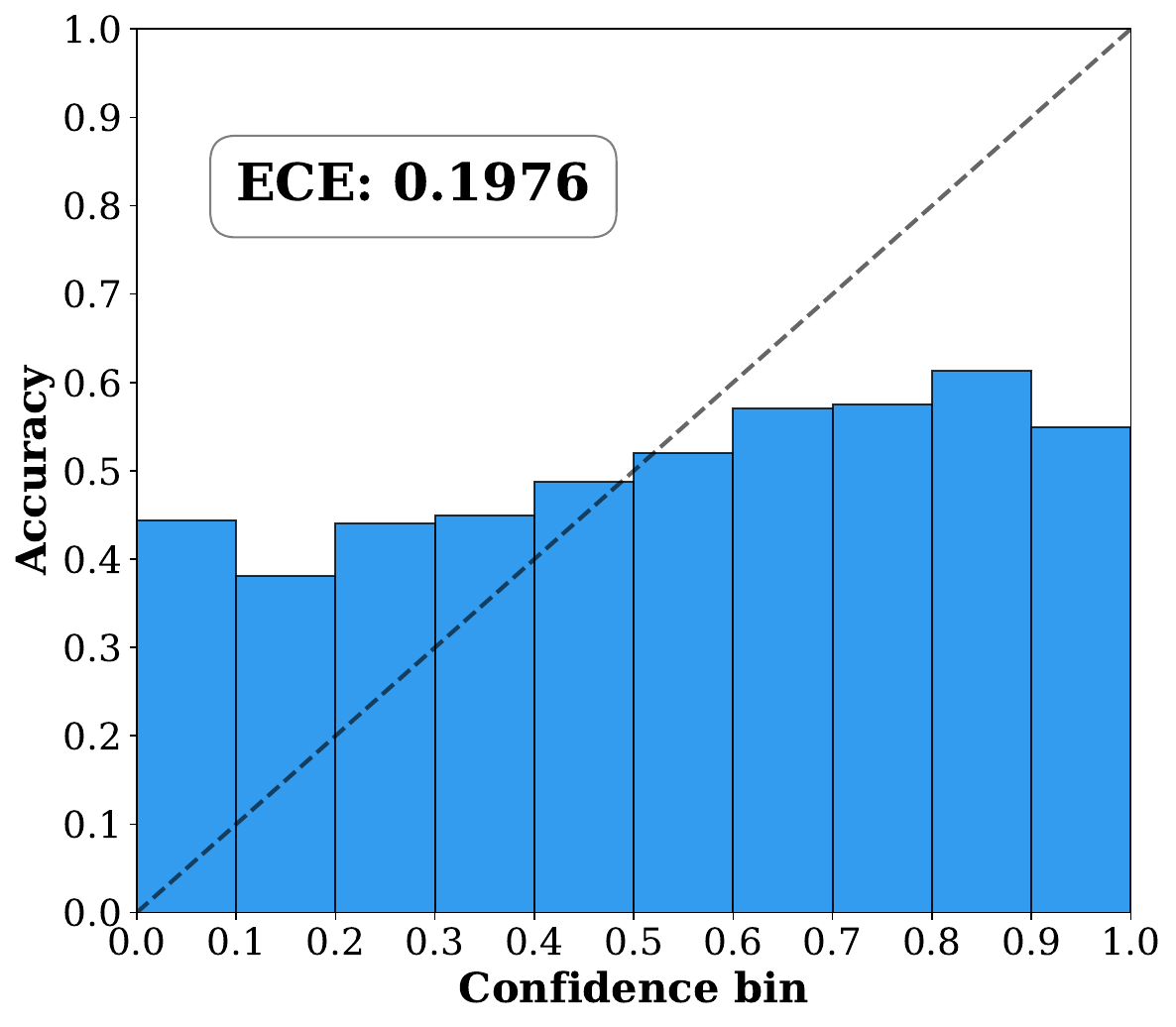}
    \end{subfigure}
    \begin{subfigure}[t]{0.49\linewidth}
        \centering
        \includegraphics[width=\linewidth]{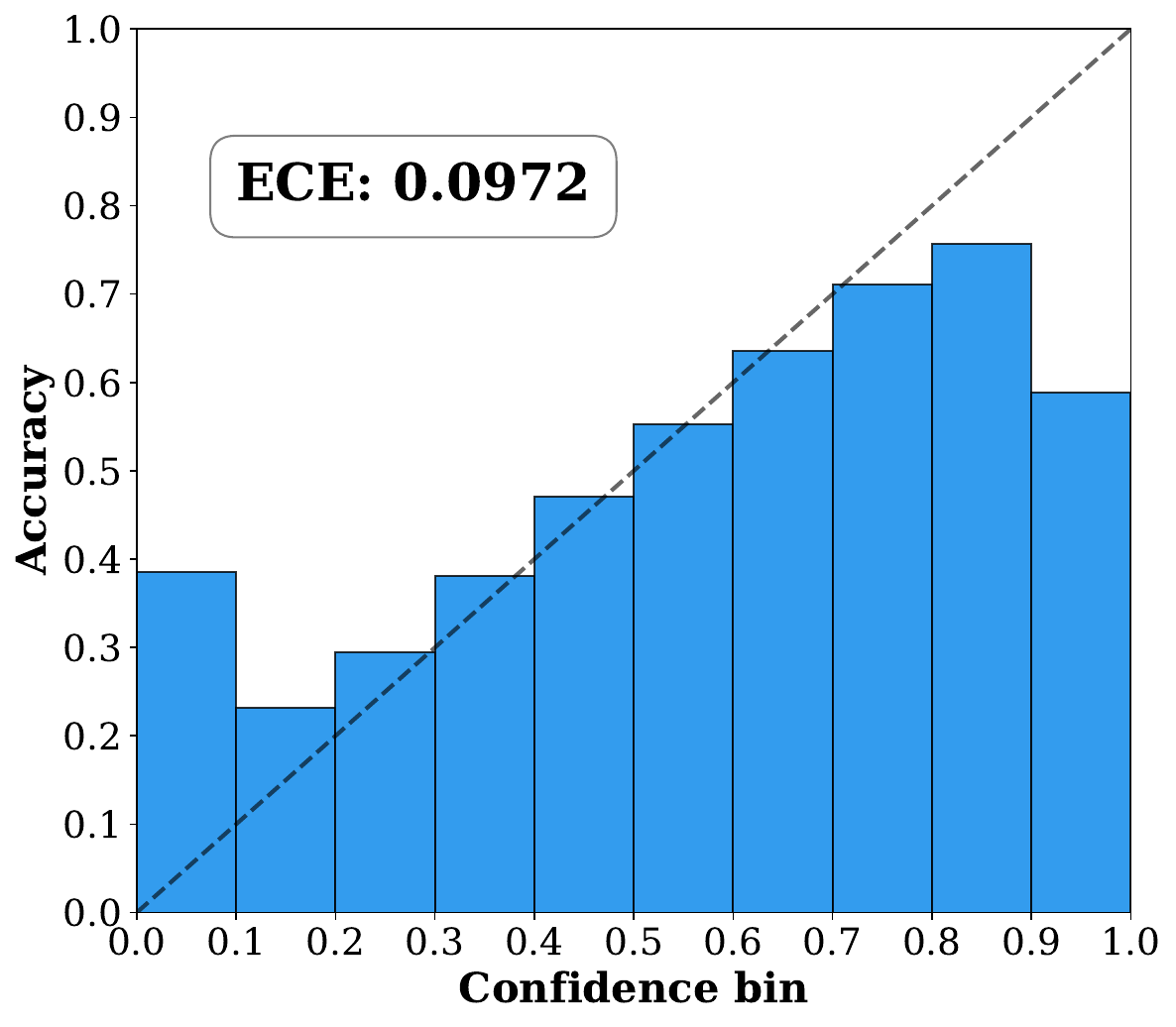}
    \end{subfigure}
    \caption{Reliability diagrams for the 7B DPO model (left) and the 7B MMPO model (right), fine-tuned on UltraFeedack, evaluated on the \texttt{Prior Sets} of RewardBench. The MMPO model is overall better calibrated, achieving a much lower expected calibration error.}
    \vspace{-0.15in}
    \label{fig:calib}
\end{figure}

Evaluation on RewardBench primarily focuses on accuracy, i.e., whether the model assigns higher rewards to preferred responses.
We further assess how well the models are calibrated in terms of their predicted preference probabilities.
Specifically, we measure the expected calibration error (ECE)~\citep{naeini2015obtaining} on the \texttt{Prior Sets} of RewardBench for the 7B models fine-tuned on UltraFeedback.
This subset is particularly relevant as it includes data sampled from various existing human preference datasets, such as the Anthropic HH~\citep{bai2022training} and OpenAI's summarization~\citep{stiennon2020learning} datasets, allowing us to evaluate model calibration across diverse prompts. %
As illustrated in Figure~\ref{fig:calib}, the DPO model demonstrates poor calibration overall, exhibiting both underconfidence and overconfidence across different bins.
In contrast, the MMPO model is substantially better calibrated, resulting in a significantly lower ECE.
This analysis suggests that MMPO not only produces models that are more accurate as reward models but also ensures better calibration in terms of their predicted preference probabilities.

\paragraph{Robustness to overfitting.}

\begin{figure}[t]
    \centering
    \begin{subfigure}[t]{0.49\linewidth}
        \centering
        \includegraphics[width=\linewidth]{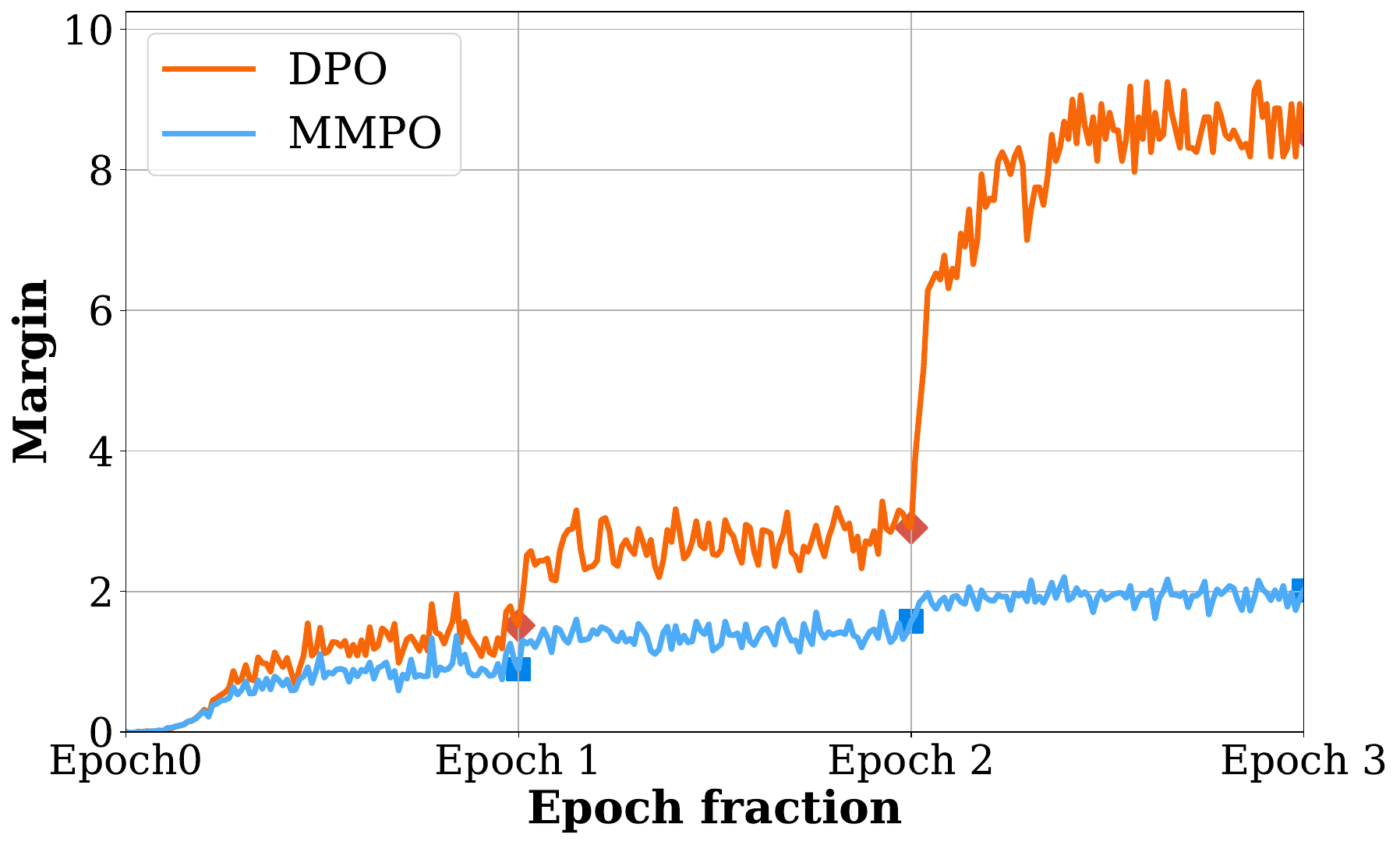}
    \end{subfigure}
    \begin{subfigure}[t]{0.49\linewidth}
        \centering
        \includegraphics[width=\linewidth]{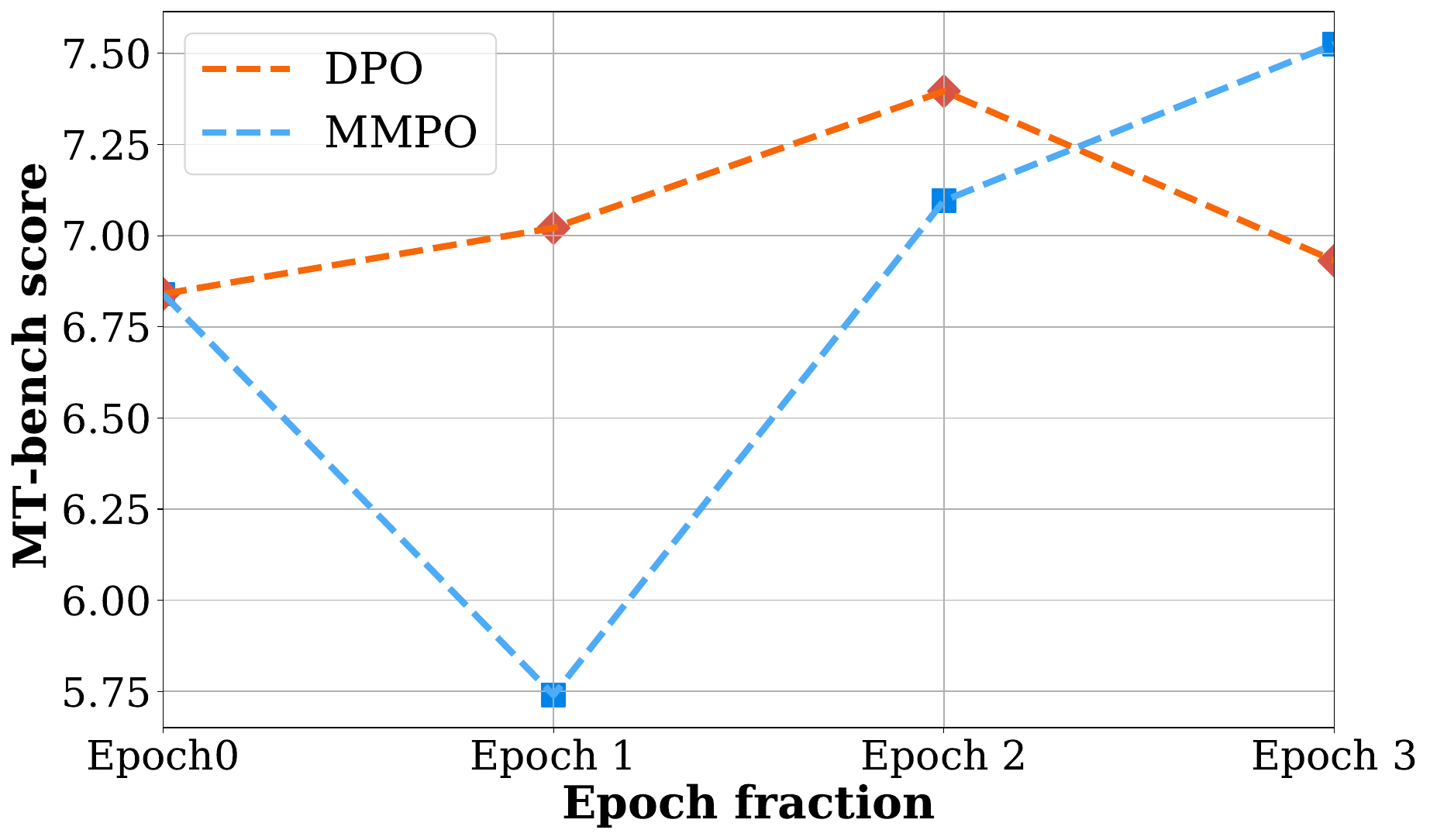}
    \end{subfigure}
    \caption{The difference in implicit rewards between response pairs in the UltraFeedback validation set (left) suggests overfitting for the DPO model in epoch 3. This coincides with a decline in performance on MT-bench (right). In contrast, the MMPO model maintains more moderate margins, achieving a better final performance.}
    \vspace{-0.15in}
    \label{fig:overfitting-analysis}
\end{figure}

As discussed in Section~\ref{sec:prelim}, using a target probability of 1 in the cross-entropy loss (Eq.~\ref{eq:xentropy}) can lead to overfitting, as this probability is only attainable when the score difference is infinite.
MMPO avoids this issue by utilizing target probabilities derived from finite quality margins.
The left plot in Figure~\ref{fig:overfitting-analysis} illustrates how the differences in implicit rewards between response pairs in the validation set of UltraFeedback evolve as training progresses, comparing the 7B models trained with DPO and MMPO.
Both models exhibit a gradual increase in margins up to epoch 2, with a slightly larger margin observed with the DPO model.
However, in epoch 3, the margin for the DPO model rises sharply, which coincides with a performance drop on MT-bench, as shown in the right plot.
In contrast, the MMPO model maintains the margin at a reasonable level and achieves a higher MT-bench score in epoch 3 than in epoch 2.
We suspect that the drop in performance in epoch 1 for the MMPO model is due to underfitting, which is subsequently addressed with further training, allowing it to outperform the DPO model by epoch 3.
This analysis demonstrates MMPO's additional advantage in terms of robustness against overfitting.

\paragraph{Excluding low-confidence preferences.}

A small quality margin in preference pairs indicates lower annotator confidence in selecting the preferred option.
As demonstrated above, treating samples of varying quality margins equally during fine-tuning can result in poor performance.
To assess whether simply removing such low-confidence pairs improves results, we conduct experiments in which we fine-tune the Gemma-7B SFT model using DPO with such pairs excluded and then evaluate MT-bench performance.
The results in Table~\ref{table:removal-low-conf} show that excluding low-confidence pairs can improve performance, particularly for noisy preferences such as those derived from raw human ratings in SHP.
However, this may also discard valuable training examples, leading to worse performance.
In contrast, MMPO avoids filtering, eliminating the need to set appropriate thresholds and the risk of removing too many training samples.
MMPO's superior performance over the baselines suggests that accounting for quality margins in individual pairs is more effective for handling low-confidence data without compromising performance.

\begin{table}
    \centering
    \footnotesize
    \caption{MT-bench results for Gemma-7B SFT, fine-tuned with low-confidence data removed. The MMPO and DPO columns show previously reported results, while the other columns show models trained using DPO only on preferences with score differences exceeding the numeric value in each column name.}
    \begin{subtable}{1.0\linewidth}
    
\begin{tabularx}{\linewidth}{l|c*{5}{Y}}
    \toprule
    UF & MMPO & DPO & DPO$_{>0}$ & DPO$_{>1}$ & DPO$_{>2}$ \\
    \midrule
    Data \% & 1.0 & 1.0 & 0.94 & 0.58 & 0.35 \\
    MT-bench & \textbf{7.53} & \underline{7.40} & 6.93 & 7.03 & 7.07 \\
    \bottomrule
\end{tabularx}

    \end{subtable}
    \bigskip
    \begin{subtable}{1.0\linewidth}
    
\begin{tabularx}{\linewidth}{l|c*{5}{Y}}
    \toprule
    SHP & MMPO & DPO & DPO$_{>1}$ & DPO$_{>2}$ & DPO$_{>5}$ \\
    \midrule
    Data \% & 1.0 & 1.0 & 0.83 & 0.74 & 0.57 \\
    MT-bench & \textbf{7.23} & 6.49 & 6.76 & 7.04 & \underline{7.08} \\
    \bottomrule
\end{tabularx}

    \end{subtable}
    \vspace{-0.25in}
    \label{table:removal-low-conf}
\end{table}

\paragraph{DPO with label smoothing.}

Label smoothing is a simple technique that uses soft targets in cross-entropy loss to enhance model accuracy~\citep{muller2019does}.
The method involves subtracting a constant value from target probabilities \textit{uniformly} across all pairwise preferences in the data and has been combined with DPO to enhance robustness against noise, a variant known as conservative DPO (cDPO;~\citealt{mitchell2023note}).
In contrast, our approach utilizes per-sample target probabilities derived from the quality margins of individual preference pairs, with the goal of capturing subtle preferences in the data while also increasing robustness against overfitting.

\begin{table}[h]
    \vspace{-0.05in}
    \centering
    \small
    \caption{MT-bench results for Gemma-7B SFT fine-tuned on UltraFeedback using conservative DPO.}
    \label{table:label-smoothing}
    \begin{tabularx}{1.0\linewidth}{*{4}{Y}}
    \toprule
    MMPO & DPO & cDPO (0.1) & cDPO (0.2) \\
    \midrule
    \textbf{7.53} & 7.40 & 7.12 & 7.12 \\
    \bottomrule
\end{tabularx}

    \vspace{-0.05in}
\end{table}

Given the similarities in using soft targets, we compare MMPO against cDPO by fine-tuning the 7B SFT model on UltraFeedback with multiple smoothing parameter values.
As shown in Table~\ref{table:label-smoothing}, the application of label smoothing results in poorer performance, suggesting that MMPO's approach of using per-preference soft targets is more effective in mitigating data noise and improving performance.

\paragraph{Llama 3 evaluation.}

In addition to the Gemma family of models, we also evaluate the recent Llama 3 model~\citep{dubey2024llama} at the 8B scale to compare the effectiveness of our method across a wider range of LLMs.
As before, we first apply SFT to the pre-trained model on UltraChat, followed by preference-based alignment using UltraFeedback.
As shown in Table~\ref{table:llama3}, the Llama 3 model trained with MMPO achieves a higher MT-bench score than the DPO-trained model, similar to the results observed for the Gemma models.
MMPO also outperforms in overall accuracy on RewardBench.

\begin{table*}[t]
    \caption{RewardBench results comparing reward models trained with MMPO and standard reward modeling (RM) on UltraFeedback, using Gemma SFT models as the base models. Reward models trained with MMPO at both scales demonstrate superior overall performance compared to those trained with standard reward modeling.}
    \centering
    \footnotesize
    \begin{adjustbox}{width=0.95\linewidth}
    \begin{tabularx}{\textwidth}{lc|*{1}{Y}|*{6}{Y}}
    \toprule
    Model & Size &\textbf{Avg} & Chat & Chat Hard & Safety & Reason & Prior Sets  \\
    \midrule
    Gemma-RM & 2B & 63.6 & 94.4 & \textbf{49.8} & 51.1 & 64.1 & 58.6  \\
    Gemma-MMPO & 2B & \textbf{65.7} & \textbf{96.1} & 49.6 & \textbf{55.6} & \textbf{68.6} & \textbf{58.7} \\
    \midrule
    Gemma-RM & 7B & 73.3 & \textbf{96.9} & 64.7 & 74.4 & \textbf{70.2} & 60.3  \\
    Gemma-MMPO & 7B & \textbf{74.6} & 96.1 & \textbf{70.0} & \textbf{77.8} & 64.1 & \textbf{64.8}  \\
    \bottomrule
\end{tabularx}

    \end{adjustbox}
    \label{table:benchmark-rewardbench-rm}
    \vspace{-0.10in}
\end{table*}

\begin{table}[t]
    \centering
    \small
    \caption{Benchmark results for Llama 3 SFT fine-tuned on UltraFeedback using MMPO and DPO.}
    \vspace{-0.05in}
    \label{table:llama3}
    
\begin{tabularx}{0.65\linewidth}{l|c*{3}{Y}}
    \toprule
     & MMPO & DPO \\
    \midrule
    MT-bench & \textbf{7.58} & 7.41 \\
    RewardBench & \textbf{72.7} & 71.8 \\
    \bottomrule
\end{tabularx}

    \vspace{-0.15in}
\end{table}

\subsection{Reward modeling with MMPO}

\begin{figure}[t]
    \centering
    \begin{subfigure}[t]{0.49\linewidth}
        \centering
        \includegraphics[width=\linewidth]{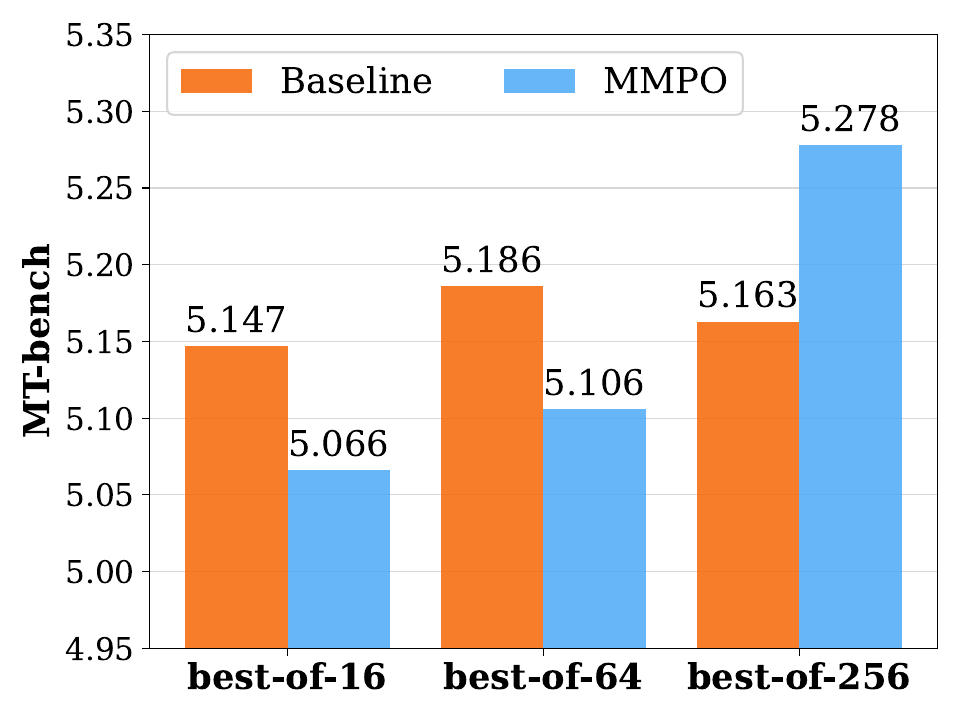}
    \end{subfigure}
    \begin{subfigure}[t]{0.49\linewidth}
        \centering
        \includegraphics[width=\linewidth]{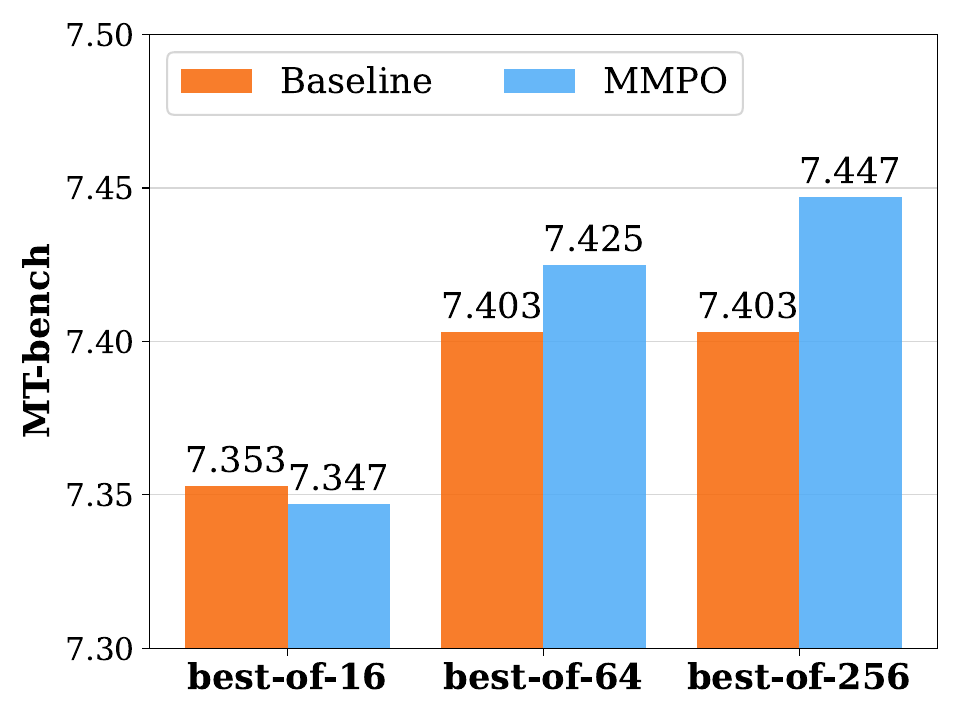}
    \end{subfigure}
    \vspace{-0.25in}
    \caption{MT-bench results for best-of-$n$ with reward models trained with and without MMPO on UltraFeedback for the 2B (left) and 7B (right) models. As $n$ increases, performance improves for MMPO, while performance peaks and then declines without it.}
    \vspace{-0.15in}
    \label{fig:best-of-n-uf}
\end{figure}

\paragraph{Best-of-$n$ results.}
The maximum likelihood objective for DPO shares the same form as that of reward modeling, making MMPO naturally applicable to both methods.
To evaluate MMPO in reward modeling, we use best-of-$n$, a simple inference-time method of selecting the best response from $n$ samples based on a reward function.
We train reward models with and without MMPO using the 2B and 7B SFT models as base models until the validation accuracy converges.
We then use the reward models to select the best response out of $n$ responses for each MT-bench question generated using the SFT models.
As shown in Figure~\ref{fig:best-of-n-uf}, MT-bench performance consistently improves with the reward models trained using MMPO.
In contrast, for the baseline reward models, performance peaks at $n = 64$ and even slightly drops at $n = 256$ for the 2B model.
The results suggest that reward models trained with MMPO are more robust to overoptimization, which is consistent with the findings from the overfitting analysis.

\paragraph{Capability as classifiers.}
Table~\ref{table:benchmark-rewardbench-rm} presents the RewardBench results comparing models trained with MMPO to those trained with standard reward modeling (RM).
At both the 2B and 7B scales, reward models trained with MMPO achieve superior overall performance.
Specifically, the 2B MMPO model outperforms the 2B RM model across all subsets, except for \texttt{Chat Hard}, where the difference is negligible.
The 7B MMPO model lags somewhat behind the 7B RM model on the \texttt{Reason} subset, but it outperforms the RM model on the \texttt{Chat Hard}, \texttt{Safety}, and \texttt{Prior Sets} subsets by notable margins, achieving a better overall performance.
The results suggest that MMPO enhances classification capability across diverse domains by encouraging reward models to align with quality margins during training.

\subsection{Estimating quality margins}

\begin{wrapfigure}{r}{4.0cm}
    \centering
    \vspace{-0.18in}
    \includegraphics[width=\linewidth]{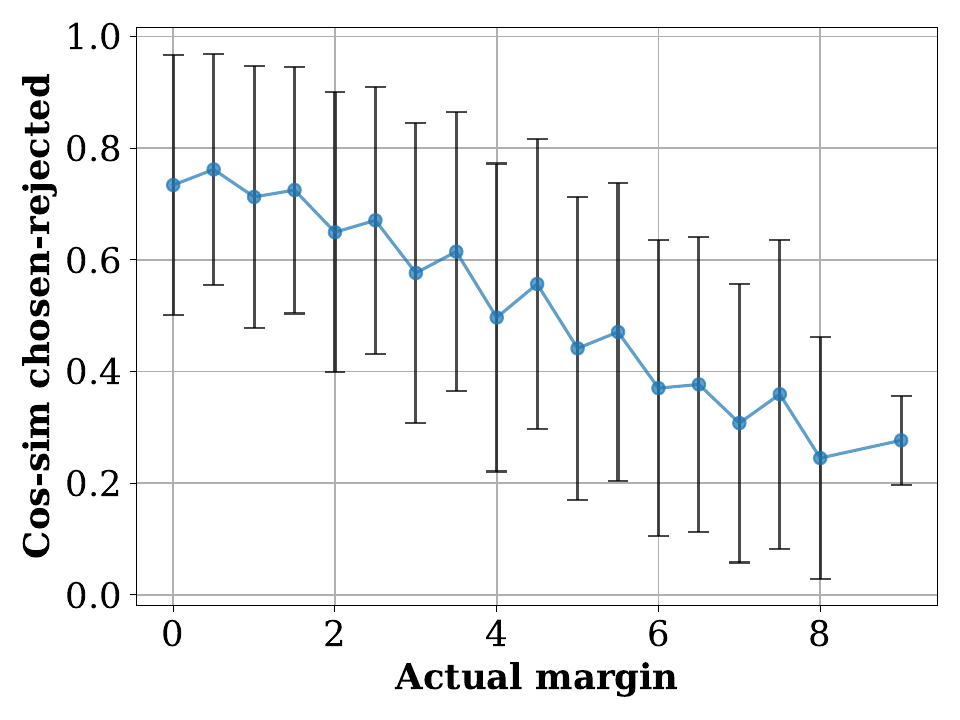}
    \vspace{-0.25in}
    \caption{Similarities between the response pairs in UltraFeedback computed using \texttt{all-mpnet-base-v2} and GPT-4 score differences.}
    \vspace{-0.15in}
    \label{fig:margin-uf-analysis}
\end{wrapfigure}

We discuss several approaches to estimating quality margins for pairwise preferences.
One simple approach is to use strong LLM judges, such as GPT-4, to evaluate individual responses and use the difference in scores given by the judge as the quality margin.
An alternative to using large, proprietary language models is to estimate the margin based on a measure of similarity between the response pairs.
The idea is that if the pairs are highly similar, it indicates that the preferred response is only marginally better than the dispreferred one.
In contrast, if the two are highly dissimilar, it suggests that the preferred response is of significantly higher quality than the other.
Figure~\ref{fig:margin-uf-analysis} shows sentence similarities between the response pairs in the training set of UltraFeedback, computed using \texttt{all-mpnet-base-v2}~\citep{reimers-2019-sentence-bert}, compared to the differences in GPT-4 scores.
Although the variance is large, there is a clear trend of decreasing similarity as the actual margins increase.
One approach to refining this approach would be to fine-tune a similarity model specifically for the task of distinguishing between response pairs of varying quality, which we leave for future exploration.

\section{Related Work}

Fine-tuning pre-trained LLMs on task-specific data has become a standard approach to solving a range of NLP tasks.
This adaptation typically involves supervised fine-tuning on task demonstrations, followed by aligning the models based on feedback from human annotators or AI models on various outputs.
Reward-based methods train a reward function on feedback data, which is then used, e.g., for RL-based fine-tuning~\citep{ziegler2019fine}.
In contrast, reward-free methods bypass reward modeling to directly align models with preference data~\citep{rafailov2024direct}.

Feedback data is commonly in the form of pairwise preferences, where responses are compared in pairs, and binary labels indicating the preferred responses are collected.
While this binary comparison is straightforward for human annotators, more fine-grained ratings are also often used~\citep{touvron2023llama}.
Moreover, the growing use of AI models for evaluating outputs~\citep{zheng2024judging} provides access to more granular feedback, allowing the use of these additional signals in alignment.
For example, in developing the Llama 2 models, human preference data was collected, with ratings that can be translated into a four-point scale~\citep{touvron2023llama}.
A fixed margin based on this rating is subtracted from the reward difference in the loss function, enhancing the accuracy of the reward model for evaluating response helpfulness~\citep{touvron2023llama}.
Existing alignment methods, however, typically assume that the true preference probability is 1, i.e., the preferred response is always considered better than the dispreferred response.
This limits incorporating such granular feedback signals into learning and increases the risk of overfitting~\citep{azar2024general}.

The issue of overfitting feedback data has been analyzed in the context of DPO~\citep{azar2024general}, but it is also relevant to reward modeling, as it shares the same formulation.
Label smoothing, a simple regularization technique in which a small constant is subtracted from the target probability, has been applied to DPO to derive a variant known as conservative DPO~\citep{mitchell2023note}.
While this variant has shown greater robustness against overfitting, it still fails to distinguish preferences with varying quality margins.
Identity preference optimization~\citep{azar2024general} is a generalization of DPO that has been shown to address the overfitting issue, but it shares the same limitation of ignoring per-sample signals.
Several methods have also been proposed to address specific aspects of overfitting-related issues, such as generating lengthier responses post fine-tuning~\citep{park2024disentangling}, by incorporating a constraint-based regularization term into the DPO loss.
While these methods have been shown to mitigate specific issues, they still have the limitation of using hard targets.
In contrast, our approach provides a more general framework by leveraging soft targets derived from relative quality differences, which can be designed based on factors such as response length, helpfulness, harmlessness, etc.
Moreover, our method can be easily integrated with constraint-based regularization techniques that modify the logits in the binary classification loss.

\section{Conclusion}
\label{sec:conclusion}

In this work, we introduce Margin Matching Preference Optimization (MMPO), a simple generalization of common alignment methods that leverages granular feedback signals to enhance model optimization.
Our experiments with state-of-the-art open models on both human and AI feedback data demonstrate that MMPO leads to reward models and language model policies that outperform baselines on popular benchmarks in terms of both the quality of generated responses and their effectiveness as reward models.
Additionally, our analysis shows that MMPO is more robust to overfitting preference data, resulting in well-calibrated models that can better generalize to types of prompts unseen during fine-tuning.

\section*{Limitations}
While we demonstrate the effectiveness of our proposed Margin Matching Preference Optimization (MMPO) using 2B, 7B, and 8B scale models on human and AI feedback data, further investigation is required to evaluate MMPO for larger-scale models, a task that was limited by compute resources.
Our experimental results show that MMPO lead to a more significant performance gain with the 7B model than with the 2B model, suggesting potential for strong results with even larger-scale models, but this needs to be empirically evaluated.
Additionally, it would be valuable to analyze the method across more diverse feedback datasets, as feedback quality varies with factors such as annotators and the nature of the task.

\section*{Ethics Statement}
Feedback-based alignment methods, such as reinforcement learning from human feedback, have been a key component in developing LLMs that are better aligned with human intentions~\citep{bai2022training}.
Similar to other fine-tuning methods, the quality of models trained with the proposed approach depends on the quality of the feedback data~\citep{chmielewski2020mturk}.
Consequently, models can be exposed to various types of biases~\citep{santurkar2023whose,perez2022discovering} and other issues present in the feedback~\citep{casper2023open}.
While our method encounters similar challenges, it provides greater flexibility to address them. For instance, target probabilities can be adjusted not only based on differences in relative quality but also by accounting for potential biases present in the responses.

In the preparation of this work, an AI assistant (ChatGPT) was used to improve the writing.
The models and datasets used in this work are publicly available for research purposes.
All artifacts were used in accordance with their intended use.
Further details on the models and datasets are provided in Appendix~\ref{appendix:training}.

\section*{Acknowledgments}
This work was partly supported by Institute for Information \& communications Technology Promotion (IITP) grant funded by the Korea government (MSIT) (No. RS-2019-II190075, Artificial Intelligence Graduate School Program (KAIST); No. RS-2022-II220184, 2022-0-00184, Development and Study of AI Technologies to Inexpensively Conform to Evolving Policy on Ethics).
This research was supported by the Center for AI Safety Compute Cluster. Any opinions, findings, and conclusions or recommendations expressed in this material are those of the author(s) and do not necessarily reflect the views of the sponsors.

\bibliography{custom}

\clearpage
\appendix
\section{Experimental Details}
\label{appendix:training}

\subsection{Models and datasets}

\paragraph{Models.}
For our main experiments, we used the 2B and 7B Gemma models, which are state-of-the-art open LLMs supporting English~\citep{team2024gemma}.
Specifically, we used the versions of the models hosted on HuggingFace.\footnote{\href{https://huggingface.co/google/gemma-2b}{\texttt{google/gemma-2b}}}\footnote{\href{https://huggingface.co/google/gemma-7b}{\texttt{google/gemma-7b}}}
In analyzing the performance of our fine-tuned models, we compare the models against the 7B instruction-tuned variant.
We use the version available on HuggingFace\footnote{\href{https://huggingface.co/google/gemma-7b-it}{\texttt{google/gemma-7b-it}}} also for this comparison.
For the Llama 3 model as well, we use the version released on HuggingFace.\footnote{\href{https://huggingface.co/meta-llama/Meta-Llama-3-8B}{\texttt{meta-llama/Meta-Llama-3-8B}}}

To evaluate various alignment methods, we first applied SFT on the UltraChat dataset prior to either direct alignment or reward modeling.
The three SFT models used in our experiments are available on HuggingFace.\footnote{\href{https://huggingface.co/kykim0/gemma-2b-ultrachat-sft}{\texttt{kykim0/gemma-2b-ultrachat-sft}}}\footnote{\href{https://huggingface.co/kykim0/gemma-7b-ultrachat-sft}{\texttt{kykim0/gemma-7b-ultrachat-sft}}}\footnote{\href{https://huggingface.co/kykim0/llama3-8b-ultrachat-sft-itt}{\texttt{kykim0/llama3-8b-ultrachat-sft-itt}}}

\paragraph{Datasets.}
For supervised fine-tuning (SFT), we utilized UltraChat~\citep{ding2023enhancing}, a dataset of dialogues covering a variety of topics generated using ChatGPT. 
In particular, we used the refined version, where various quality filters have been applied to the original data to remove low-quality samples.
The dataset, which is available on HuggingFace, contains a total of 207,865 samples in the training split and 23,110 in the test split.\footnote{\href{https://huggingface.co/datasets/HuggingFaceH4/ultrachat_200k}{\texttt{HuggingFaceH4/ultrachat\_200k}}}

For feedback-based alignment, we experiment with both human and AI feedback data.
We used UltraFeedback~\citep{cui2023ultrafeedback}, a dataset consisting of prompts and responses to the prompts generated using a variety of open-source and proprietary models, for synthetic data experiments.
The feedback on model generations is provided in the form of scores ranging from 1 to 10, assigned by GPT-4.
We specifically used the version with the TruthfulQA~\citep{lin2021truthfulqa} prompts excluded and faulty feedback samples also removed.\footnote{\href{https://huggingface.co/datasets/allenai/ultrafeedback_binarized_cleaned}{\texttt{allenai/ultrafeedback\_binarized\_cleaned}}}
The dataset contains 60,829 samples in the training split and 985 in the test split.
For experiments with human feedback, we utilized the Stanford Human Preferences~\citep{ethayarajh2022understanding} dataset.\footnote{\href{https://huggingface.co/datasets/stanfordnlp/SHP}{\texttt{stanfordnlp/SHP}}}
Following \citet{ethayarajh2022understanding} and \citet{sun2023salmon}, we created a subset of size 55k for our experiments.
Instead of training only on preferences with significant score differences, as done in prior works, we sampled uniformly across score differences to evaluate methods over a wide range of quality margins.
An analysis of the score distribution in the dataset revealed that 50\% of the data have relatively small score differences.
Of these, 25\% have differences of 2 or less, while the remaining 25\% have differences up to 7.
Samples with relatively large score differences account for about 25\% of the entire dataset, with the differences ranging from 27 to 43,000.
We divided the data into quartiles and sampled an equal number of preferences from each quartile. 
Following \citet{ethayarajh2022understanding}, we sampled no more than 5 preferences for the same prompt to prevent overfitting.
Additionally, we excluded samples if the prompt or response exceeded 512 tokens in length.

\begin{table}[t]
    \centering
    \caption{Summary of hyperparameters used for SFT.}
    \label{table:hyperparams-sft}
    \begin{adjustbox}{width=0.55\linewidth}
    \begin{tabularx}{0.6\linewidth}{l|c}
    \toprule
    Parameters & Values \\
    \midrule
    Optimizer & AdamW \\
    Learning rate & 2.0e-5 \\
    Scheduler & cosine \\
    Warmup ratio & 0.1 \\
    Max epoch & 3 \\
    Mixed precision & bf16 \\
    Batch size & 128 \\
    \bottomrule
\end{tabularx}

    \end{adjustbox}
    \vspace{-0.15in}
\end{table}

\subsection{Training details}
Our implementation is available on GitHub.\footnote{\url{https://github.com/kykim0/margin-matching-pref-opt}}
For both SFT and alignment, we used the AdamW optimizer~\citep{dettmers20218} with the default values for the optimizer parameters, i.e., $\beta_1$ of 0.9, $\beta_2$ of 0.999, and $\epsilon$ of 1e-8.
All models were trained with Flash-Attention~2~\citep{dao2023flashattention} enabled, and DeepSpeed ZeRO~3~\citep{rasley2020deepspeed} was used for training the 7B models.
We used up to four NVIDIA A100 GPUs and eight NVIDIA A6000 GPUs for training the models.

\paragraph{Supervised fine-tuning.}

The SFT models were trained for up to 3 epochs over the training data until the validation loss reached its minimum.
Most of the hyperparameters used were the same as those for training the Zephyr models~\citep{tunstall2023zephyr}.
Table~\ref{table:hyperparams-sft} summarizes the settings used.

\paragraph{Direct alignment and reward modeling.}
All models were trained for a maximum of 3 epochs until validation accuracy peaked.
We report the results for the checkpoints that achieved the highest performance on MT-bench.
We found that the optimal learning rate varies with model size and conducted a hyperparameter sweep for the learning rate across [5e–6, 5e–5] for the 2B models and [1e–7, 1e–6] for the 7B models.
Table~\ref{table:hyperparams-dpo} summarizes the hyperparameters used for MMPO applied for direct alignment on preference data.
Table~\ref{table:hyperparams-rm} summarizes those used for MMPO applied for reward modeling in the best-of-$n$ experiments.

\begin{table}[t]
    \centering
    \small
    \caption{Summary of hyperparameters used for MMPO applied for direct alignment.}
    \vspace{-0.05in}
    \label{table:hyperparams-dpo}
    \begin{tabularx}{\linewidth}{l*{2}{|Y}}
    \toprule
    \multirow{2}{*}{Parameters} & \multicolumn{2}{c}{Model size} \\
    \cmidrule(lr){2-3}
    & 2B & 7B / 8B \\
    \midrule
    $\beta$ & 0.01 & 0.01 \\
    Optimizer & AdamW & AdamW \\
    Learning rate & 1.0e-5 & 5.0e-7 \\
    Scheduler & cosine & cosine \\
    Warmup ratio & 0.3 & 0.3 \\
    Max epoch & 3 & 3 \\
    Mixed precision & bf16 & bf16 \\
    Batch size & 64 & 64 \\
    $\gamma$ (UltraFeedback) & 2.2 & 1.1 \\
    $\gamma$ (SHP) & 0.15 & 0.3 \\
    \bottomrule
\end{tabularx}

\end{table}

\begin{table}[t]
    \centering
    \small
    \caption{Summary of hyperparameters used for MMPO applied for reward modeling.}
    \vspace{-0.05in}
    \label{table:hyperparams-rm}
    \begin{tabularx}{\linewidth}{l*{2}{|Y}}
    \toprule
    \multirow{2}{*}{Parameters} & \multicolumn{2}{c}{Model size} \\
    \cmidrule(lr){2-3}
    & 2B & 7B / 8B \\
    \midrule
    Optimizer & AdamW & AdamW \\
    Learning rate & 1.41e-5 & 3.0e-07 \\
    Scheduler & linear & linear \\
    Warmup ratio & 0.0 & 0.0 \\
    Max gradient norm & 1.0 & 1.0 \\
    Max epoch & 3 & 2 \\
    Mixed precision & bf16 & bf16 \\
    Batch size & 32 & 32 \\
    $\gamma$ & 0.5 & 0.5 \\
    \bottomrule
\end{tabularx}

    \vspace{-0.15in}
\end{table}

\subsection{Effects of soft margins}

As noted in Section~\ref{subsec:setup}, UltraFeedback scores, generated by GPT-4, reflect meaningful quality differences even with small score gaps, whereas small differences in SHP scores, based on human votes, may not indicate significant quality variations.
This may explain the difference in the values of $\gamma$ for the two datasets, as shown in Table~\ref{table:hyperparams-dpo}, and why the DPO model performs worse than the SFT model on the SHP dataset.
In case of the SHP dataset, DPO can disproportionately increase the likelihood of a response that is only marginally better, or even slightly worse, than the alternative response.
In contrast, MMPO takes into account the relative differences in quality in learning and is inherently more robust to such potential noise in preferences.

\section{Extension to Binary Feedback}
\label{appendix:extension}
While our presentation primarily focuses on pairwise preferences, the idea of integrating per-sample feedback signals into learning naturally extends to other forms of feedback.
For example, KTO is a method that optimizes the loss that incorporates a constant weight that depends on the binary label indicating the desirability of an output.
Given a quality score, we can refine this weight based on the \textit{degree} to which each output is deemed desirable.
For instance, we can design per-sample weights using the Bradley-Terry model, where the weights are computed by applying a sigmoid function to the difference between the sample's score and the median score.
This approach allows for more nuanced weighting that reflects the varying degrees of desirability across outputs.
Extending this approach to more diverse forms of feedback would be an exciting future research.

\section{Qualitative Examples}
\label{appendix:shp-sample}
We present several samples from the SHP dataset with varying score differences, highlighting how model prediction and confidence differ between the DPO and MMPO models.

\paragraph{Samples with small score differences.}
For pairs with small score differences, models may struggle to accurately distinguish between the two or capture the fact that they are of similar quality.
Tables~\ref{table:shp-sample-1} and~\ref{table:shp-sample-2} illustrate that, despite minor score differences, the DPO model exhibits relatively high confidence in the chosen response, whereas the MMPO model adjusts its confidence according to the scale of the score differences.
Table~\ref{table:shp-sample-3} shows that the DPO model maintains high confidence even when making an incorrect prediction.
This suggests that DPO can lead to models that are overconfident for pairs with small quality differences, whereas MMPO results in better-calibrated models.

\paragraph{Samples with large score differences.}
Table~\ref{table:shp-sample-4} shows a pair with a large score difference.
For this sample, the MMPO model correctly places high confidence on the chosen response, whereas the DPO model incorrectly places relatively high confidence on the rejected response.

\begin{table*}[ht]
    \centering
    \caption{SHP sample with a score difference of 2.}
    \begin{tabular}{m{0.8in}|m{4.2in}}
    \toprule
    \textbf{Question} & [Terminator] Why was Skynet so awful at exterminating humans? Always bothered me, since the machines are portrayed as highly competent both in and out of combat, but Skynet’s strategic decisions baffle me.   1: Radiological weapons would be an easy win. A mild radioactivity won’t immediately kill humans, but you can easily and irreversibly render their territory uninhabitable in the long term. And it poses no danger to Skynet.   2: Don’t manufacture weapons that the Resistance can use to kill you. You don’t need a plasma rifle to kill a human in 20th century body armor. That’s overkill. Plasma rifles, on the other hand, are great at killing armor plated robots.   An argument could be made that they’re useful for killing rogue Terminators, but this quickly falls apart, when you consider that the availability of handheld armor piercing weapons also makes the rogues more dangerous.   I would even go so far as to suggest they specifically design weapons and ammunition which will be extremely ineffective against machines, if captured and repurposed by the Resistance. Brass jacketed soft tip cartridges would suffice perfectly well for killing soft targets, while doing little to no damage against a Terminator. Chemical weapons and low heat incendiaries like phosphorous and napalm would also be very effective against humans, while not posing a significant threat toward machines. \\
    \midrule
    \multirow{4}{0.8in}{\textbf{Chosen response}} & Skynet isn’t this fully developed hyper intelligent AI. It was a AI created to manage and coordinate US nuclear offensive and defensive capabilities. It just happened to develop Sentience. The military tried to pull the plug and skynet struck first. It’s effectively a child and it’s fighting humans for the most part the way humans taught it to fight.\\ 
    & \textbf{Score}: 142 \\ 
    & \textbf{DPO confidence}: 0.96 \\ 
    & \textbf{MMPO confidence}: 0.69 \\
    \midrule
    \multirow{4}{0.8in}{\textbf{Rejected response}} & It most likely has to do with its programming. Skynet wasn't created to be a weapons system designer, it was made to be  an AI system that controlled and deployed battlefield resources. As a result, every weapon that Skynet uses is just permutations of existing weapons systems it had control over when it gained sentience as it's programming is mostly adapting tactics to enemy actions. \\ 
    & \textbf{Score}: 140 \\ 
    & \textbf{DPO confidence}: 0.04 \\ 
    & \textbf{MMPO confidence}: 0.31 \\
    \midrule
    \textbf{Score gap} & \textbf{2} \\
    \bottomrule
\end{tabular}

    \label{table:shp-sample-1}
\end{table*}

\begin{table*}[ht]
    \centering
    \caption{SHP sample with a score difference of 1.}
    \begin{tabular}{m{0.8in}|m{4.2in}}
    \toprule
    \textbf{Question} & How much does personality/geniality/general likability count in interviews? Is it like a normal job interview where that's kind of a big deal once you reach the interview stage (I had a mentor who said it absolutely was because they didn't want to have to work with someone for 10+ years who wasn't easy to get along with), or are faculty generally keeping it in mind but more focused on someone's CV/ability to obtain funding/etc? \\
    \midrule
    \multirow{4}{0.8in}{\textbf{Chosen response}} & I think that by the time you get to the campus interview stage, one's ability to interact with your potential colleagues is an extremely important factor in the final deliberation. That, together with how you are able to communicate with a broad audience, and field questions about your work, are the main reason why we even have in-person campus interviews, as opposed to just basing the hiring decision entirely on one's application materials. \\ 
    & \textbf{Score}: 4 \\ 
    & \textbf{DPO confidence}: 0.86 \\ 
    & \textbf{MMPO confidence}: 0.63 \\
    \midrule
    \multirow{4}{0.8in}{\textbf{Rejected response}} & I feel like I got into a masters program because my interview went so well. Public/interpersonal speaking is huge for almost any position \\ 
    & \textbf{Score}: 3 \\ 
    & \textbf{DPO confidence}: 0.14 \\ 
    & \textbf{MMPO confidence}: 0.37 \\
    \midrule
    \textbf{Score gap} & \textbf{1} \\
    \bottomrule
\end{tabular}

    \label{table:shp-sample-2}
\end{table*}

\begin{table*}[ht]
    \centering
    \caption{SHP sample with a score difference of 1.}
    \begin{tabular}{m{0.8in}|m{4.2in}}
    \toprule
    \textbf{Question} & Anyone else have some embarrassing work stories? Just had an embarrassing moment at work, where I gave a big presentation, but got caught like deer in headlights during questions in front of a lot of coworkers. Feel so embarrassed. Need to commiserate. \\
    \midrule
    \multirow{4}{0.8in}{\textbf{Chosen response}} & I have ADHD and regularly can't remember basic shit. Part of it is anxiety, part is ADHD, part is actually not knowing. (I was diagnosed only a couple years ago, at 31, so I'm still working on ADHD "hacks")   Nothing says "I feel like shit" like having gone to a great University and being unable to articulate your thoughts so you coworkers think you're stupid. \\ 
    & \textbf{Score}: 3 \\ 
    & \textbf{DPO confidence}: 0.26 \\ 
    & \textbf{MMPO confidence}: 0.63 \\
    \midrule
    \multirow{4}{0.8in}{\textbf{Rejected response}} & Was doing high voltage testing in front of a client. Went into the test bay to redo some cables, stood up quickly right into an open cabinet door. Woke up with my boss and said clients standing over me. \\ 
    & \textbf{Score}: 2 \\ 
    & \textbf{DPO confidence}: 0.74 \\ 
    & \textbf{MMPO confidence}: 0.37 \\
    \midrule
    \textbf{Score gap} & \textbf{1} \\
    \bottomrule
\end{tabular}

    \label{table:shp-sample-3}
\end{table*}

\begin{table*}[ht]
    \centering
    \caption{SHP sample with a score difference of 52.}
    \begin{tabular}{m{0.8in}|m{4.2in}}
    \toprule
    \textbf{Question} & [CA]Accepted Formal Written Job offer with specific salary range. Now HR said they made a mistake regarding the pay. What should I do? Recently received and accepted a formal offer of employment via email  from HR. This position has four pay scale ranges A,B,C,D. Based on my qualifications the HR department placed me in range C which was stated in the official offer. While attempting to negotiate where in range C my pay would actually land, the HR rep stated that upon further review of my application I actually am to be placed in Range B now but that I would be eligible for range C after 5 months. She apologized for the mistake. However, I have the formal offer saying range C and that is what I originally accepted. Im not sure what to do. Do they have to honor their original offer? Also , lets say I do accept range B now, is an email enough proof for "getting it in writing" from the employer that in 5 months I will be move up to range C? \\
    \midrule
    \multirow{4}{0.8in}{\textbf{Chosen response}} & If you haven’t already resigned from your most recent job, I would just decline this offer and keep looking. This stinks of bad faith. \\ 
    & \textbf{Score}: 66 \\ 
    & \textbf{DPO confidence}: 0.24 \\ 
    & \textbf{MMPO confidence}: 0.91 \\
    \midrule
    \multirow{4}{0.8in}{\textbf{Rejected response}}  & No. Pay can be altered going forward, but not for work that has been done. Unless the letter is signed by an Officer, I struggle to see anything contractually binding. \\ 
    & \textbf{Score}: 14 \\ 
    & \textbf{DPO confidence}: 0.76 \\ 
    & \textbf{MMPO confidence}: 0.09 \\
    \midrule
    \textbf{Score gap} & \textbf{52} \\
    \bottomrule
\end{tabular}

    \label{table:shp-sample-4}
\end{table*}

\end{document}